\def\year{2022}\relax
\documentclass[letterpaper]{article} 
\usepackage{aaai22}  
\usepackage{times}  
\usepackage{helvet}  
\usepackage{courier}  
\usepackage[hyphens]{url}  
\usepackage{graphicx} 
\usepackage{natbib}  
\usepackage{caption} 
\DeclareCaptionStyle{ruled}{labelfont=normalfont,labelsep=colon,strut=off} 
\usepackage{algorithm}
\usepackage{algpseudocode}  
\usepackage{newfloat}
\usepackage{listings}
\floatstyle{ruled}
\newfloat{listing}{tb}{lst}{}
\floatname{listing}{Listing}

\usepackage{verbatim}
\usepackage{times}
\usepackage{amsmath}
\usepackage{multirow}
\usepackage{multicol}
\usepackage{kotex}
\usepackage{amssymb}
\usepackage{pifont}
\usepackage{algpseudocode}
\usepackage{subfigure}
\usepackage{subfig}
\usepackage{epstopdf}
\usepackage{array}
\usepackage{booktabs}

\title{NaturalInversion: Data-Free Image Synthesis Improving Real-World Consistency}
\author{
    Yujin Kim\textsuperscript{\rm 1,2}, Dogyun Park\textsuperscript{\rm 1,2}, Dohee Kim\textsuperscript{\rm 2}, Suhyun Kim\textsuperscript{\rm 2}\thanks{Corresponding author.}
}
\affiliations{
    \textsuperscript{\rm 1}Korea University, Republic of Korea\\
    \textsuperscript{\rm 2}Korea Institute of Science and Technology, Republic of Korea\\
    \{lakeeye1220, koparkrea, kimdohe1070, dr.suhyun.kim\}@gmail.com
    
}

\begin{document}

\maketitle

\begin{abstract}

We introduce \textit{NaturalInversion}, a novel model inversion-based method to synthesize images that agrees well with the original data distribution without using real data. In NaturalInversion, we propose: (1) a \textit{Feature Transfer Pyramid} which uses enhanced image prior of the original data by combining the multi-scale feature maps extracted from the pre-trained classifier, (2) a \textit{one-to-one} approach generative model where only one batch of images are synthesized by one generator to bring the non-linearity to optimization and to ease the overall optimizing process,
(3) learnable \textit{Adaptive Channel Scaling} parameters which are end-to-end trained to scale the output image channel to utilize the original image prior further. 
With our NaturalInversion, we synthesize images from classifiers trained on CIFAR-10/100 and show that our images are more consistent with original data distribution than prior works by visualization and additional analysis. Furthermore, our synthesized images outperform prior works on various applications such as knowledge distillation and pruning, demonstrating the effectiveness of our proposed method. Code is available at \url{https://github.com/kdst-team/NaturalInversion.git}
\end{abstract}

\begin{figure*}[t]
\centering
    \includegraphics[width=0.92\textwidth]{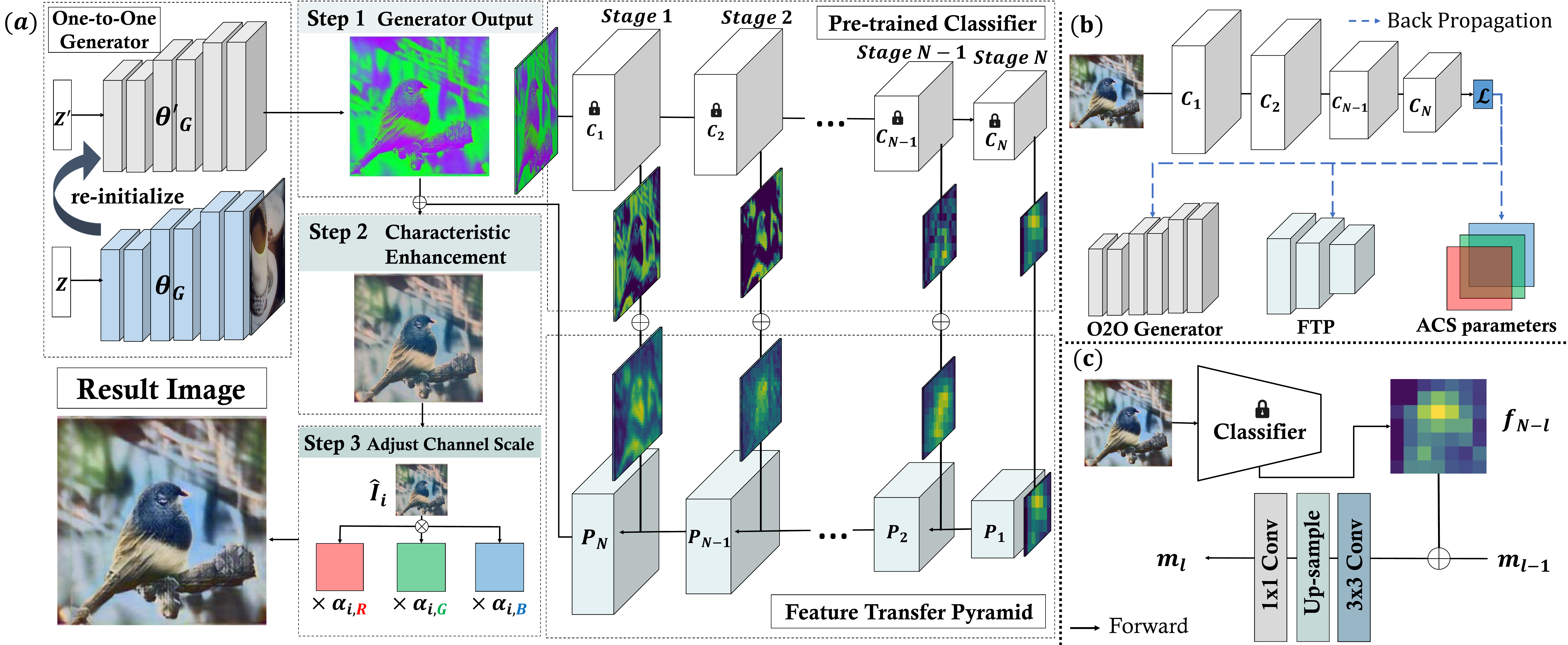}
    \hfil
\caption{(a) Overall diagram of NaturalInversion for synthesizing high-quality images given pre-trained classifier. NaturalInversion pipeline consists of 3 parts : Step1. The generator produces one batch of samples and re-initializes after the current batch synthesis step. Step 2. Feature Transfer Pyramid (FTP) works in full conjunction with a pre-trained classifier, extracting feature maps and enhancing the characteristics of the original data. Step 3. Multiply the Adaptive Channel Scaling parameters to Step2 images, and then we obtain final images. (b) Backpropagation phase to feed final images to a pre-trained classifier and update all components. (c) Illustration of a single FTP block.} 
\label{Figure1}
\end{figure*}

\section{Introduction}
\noindent Convolution Neural Networks (CNNs) have achieved a great success in various computer vision tasks such as image classification, image generation, etc \cite{simonyan2014very,goodfellow2014generative}. The emergence of large datasets \cite{Krizhevsky_2009_17719,deng2009imagenet} has led to a progressive improvement in various applications \cite{zhai2021scaling,chu2020detection}. The CNNs learn meaningful feature spaces with rich information from low-level features to high-level semantic content \cite{zeiler2014visualizing}.
Hence, several works have been proposed to transfer knowledge 
from pre-trained networks into a lightweight model to inference in various conditions. Specially, Knowledge Distillation \cite{hinton2015distilling} transfers the knowledge from the pre-trained teacher network to student networks. Network pruning \cite{liu2017learning} reduces redundant neural connections of a pre-trained network and fine-tunes the remaining weights to recover the accuracy.
However, these methods require original data to train or fine-tune the compressed networks, limiting their use in privacy-sensitive or data-limited scenarios.

To overcome these scenarios, several approaches have synthesized the data from a pre-trained network without original datasets or images prior \cite{mahendran2015understanding}. 
Recently, \citet{yin2020dreaming} has proposed a novel method to optimize the raw input space $\hat{x}$ to synthesize images, using a regularizer that follows the statistics in the batch-normalization(BN) layers of a pre-trained classifier. 
However, we argue that optimizing the high-dimensional raw input space is challenging for two reasons: 1) limited image prior since the statistics in BN layer are averaged, ignoring the specific information for individual images, 2) optimization without non-linearity. These problems lead to sub-optimized images which are heavily inconsistent with original data distribution: low fidelity and low diversity of synthesized images. Therefore, these images cause performance degradation in applications compared to performance trained with original dataset.

Therefore, for the richer image prior, we focus on features of pre-trained CNNs, since they encode the low to high level image prior on multi-scale feature maps \cite{islam2020much}. The sparsity in feature maps eliminates the irrelevant variability of the input data while preserving the important information \cite{xu2021generative}. Furthermore, feature maps of each layer encourage to capture multi-scale characteristics of the target class and real data distribution. Thus, utilizing the feature maps of the pre-trained classifier for generation strengthens the real data characteristics on the synthesized images \cite{shocher2020semantic,wang2021imagine}.


Inspired by the above methods, we propose the sub-generator: \textit{Feature Transfer Pyramid} (FTP) that uses multi-scale feature maps from the pre-trained network to generated images. Thus, FTP enhances the characteristics of real data by sequentially combining the multi-scale feature maps from a pre-trained classifier. 
By adding the final combined multi-scale feature maps to synthesized image, it strengthens the original data characteristics on the synthesized images, making images more consistent to original data distribution. Secondly, we use the CNN-based generator to bring a non-linearity in optimization, which makes generator possess higher ability to represent the complex characteristics. We further propose the \textit{one-to-one} approach, which utilizes a generator for one batch of images during the current batch synthesis step. 
Specifically, each generator is optimized to generate a specific subset of the original datasets, which accelerate the convergence to the specific optimal point. In addition, we use the learnable \textit{Adaptive Channel Scaling} (ACS) parameters that are end-to-end trained to adjust the output channel scale to find ``optimal channel scale range" of original data learned by the classifier. Thus, ACS parameters implicitly learn the image prior.

In the end, with our proposed method called \textit{NaturalInversion}, we synthesize more optimized images which agrees more with original data distribution: higher fidelity and diversity than prior works.
Experiments on the CIFAR dataset show that NaturalInversion not only synthesize more visually plausible samples than prior works but also achieves significantly higher performance on empirical studies and applications in data-free conditions. The contributions of our proposed methods can be summarized as follows: 
\begin{itemize}
\item We propose a \textit{Feature Transfer Pyramid}, which enhances the synthesized images with the multi-scale feature maps extracted from a pre-trained classifier.
\item We introduce the \textit{one-to-one} approach based generator to ease the optimizing process and bring a non-linear optimization.
\item We further implicitly learn the image prior of original data channel scale by end-to-end training the \textit{Adaptive Channel Scaling} parameters.
\end{itemize}

\section{Related Work}
 \paragraph{Model Inversion.}Inverting a pre-trained model helps to interpret and understand the deep representation that the model stored. Several works have reconstructed what the model learned by optimizing input space from noise to image with regularizers. DeepDream \cite{mordvintsev2015deepdream} visualizes the pattern that the model watch by selecting one or more layers of the model and optimizing the input space to maximize the class probability using the arbitrary target label $\hat{y}$. Follow-up studies such as
 DAFL \cite{chen2019data} simultaneously perform inversion and knowledge distillation, training a generator to minimize the cross-entropy loss and maximize the representation of the pre-trained teacher network. DFAD \cite{fang2019data} trains the generator that maximizes disagreement between the predictions of the pre-trained teacher and the student model, while a student network reduces the discrepancy with a teacher to generate more confusing samples. However, DAFL and DFAD concentrate on synthesizing the images for a specific task, such as knowledge distillation. Thus, these images are far from the original data distribution, leading to
 performance degradation on various vision applications. To alleviate this issue, \citet{yin2020dreaming,haroush2020knowledge} use a regularizer for channel-wise feature distribution matching with stored BN statistics to improve consistency with original data distribution, achieving the improvement in other applications. However, due to the problems with optimizing the input space with limited image prior, their images still lack of fidelity and diversity.
\paragraph{Usage of Feature Maps.}
The feature maps from pre-trained CNNs can provide insight into which characteristic in feature map does the classifier intensively detects for.
The low-level feature maps generated by the shallower layers of CNNs encode basic representations such as edges and corners of real images. The deeper features include more complex semantic representations such as complicated geometric shapes in their feature maps \cite{yosinski2015understanding}. For confirming the encoded information of CNNs, many works visualize feature maps of a pre-trained model to interpret feature representation in intermediate layers \cite{zeiler2014visualizing,olah2017feature}. Inspired by previous works, several studies have synthesized realistic images using the properties of feature maps \cite{kalischek2021light,heitz2021sliced,xu2021generative}. Many style transfer algorithms \cite{gatys2016image,lin2021drafting,kalischek2021light} extract the multi-scale feature maps on multiple layers by forwarding the style image to the network. Then they render the style from an image to a content image while preserving the semantic information of the content image. \citet{gatys2015texture} utilizes the correlations between feature maps in several layers to capture spatial-based texture representations. GAN-based Semantic Generation Pyramid \cite{shocher2020semantic} proposes to feed the target image to a pre-trained classifier and combine both the feature map and same level generator block output to replicate the classifier features.  
We draw inspiration from several tasks of using multi-scale feature maps to obtain different scale representations of pre-trained CNNs. 
Our proposed method gradually enhances the real data characteristics encapsulated in a pre-trained teacher network to 
capture the distribution of the original dataset.

\begin{figure*}[t]
\centering
    \includegraphics[width=0.7\textwidth,height=0.5\columnwidth]{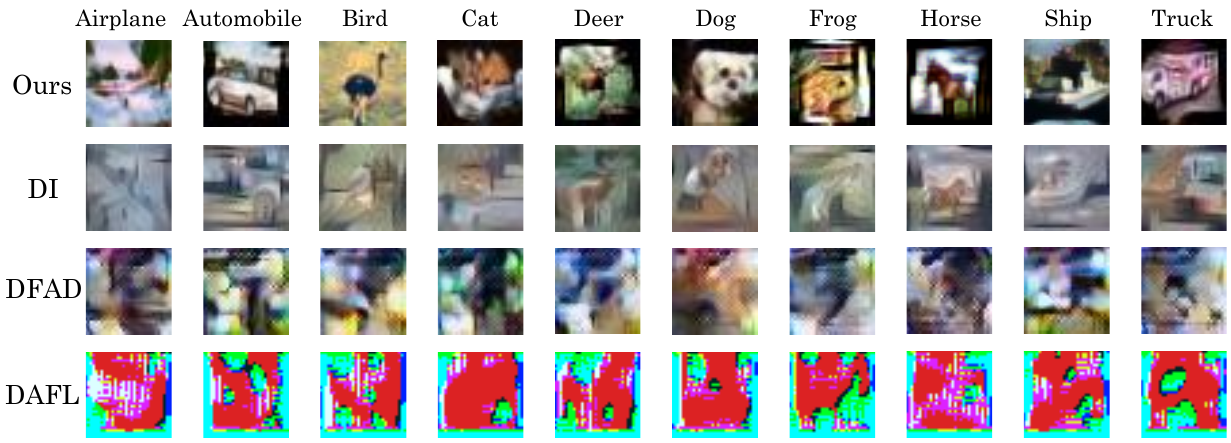}
    \hfil
\caption{Visualization of CIFAR-10 examples compare to prior works. From bottom to up, the result images were synthesized with DAFL, DFAD, DI, and our method, respectively. All images are synthesized by the same pre-trained classifier, ResNet-34. The CIFAR-10 class labels are written at the top.}
\label{Figure2}
\end{figure*}
\section{Method}
We aim to synthesize more realistic images which agrees well with the original data distribution.
In this section, we provide a brief background and introduce our \textit{NaturalInversion} methods.
\subsection{Preliminary}
Model inversion approaches update the input space to get an image $\hat{x}$ given the pre-trained teacher network $T$. Given trainable input space $\hat{x}$ with randomly initialized noise and arbitrary target label $y$, the input space is updated by minimizing the below objective function.
\begin{equation}\label{objective}
\min\limits_{\hat{x}} \mathcal{L} (\hat{x},y) + \mathcal{R}(\hat{x})
\end{equation}
We can divide objective function into two part, loss function $\mathcal{L}$ and regularizer $\mathcal{R}$ that captures natural image prior. 

\paragraph{Inception Loss.} Inceptionism-style images synthesis approach \cite{mordvintsev2015deepdream} uses inception loss to maximizes the probability of the expected target class produced by the pre-trained teacher model. Given an arbitrary target label $\hat{y}$, it encourages minimizing the cross-entropy loss to find out the class probability distribution of original data, as below:
\begin{equation}\label{eq2}
\mathcal{L}_{CE} =  CE(T(\hat{x}),{y} ) = -\sum\limits_{i}\hat{y_i} \log y_i
\end{equation}
\paragraph{Feature Distribution Regularizer.} For more image prior, DI \cite{yin2020dreaming} synthesizes the images that follow the original dataset distribution stored in the pre-trained teacher network. Given running mean ($\hat\mu$), and running variance($\hat\sigma^2$) of each BN layer \cite{ioffe2015batch}, they minimize the difference between the channel-wise mean($\mu(\hat{x})$) and variance($\sigma^2(\hat{x})$) of synthesized images and running statistics of every BN layer. 
\begin{equation}\label{eq3}
\mathcal{R}_{BN} = \sum\limits_{i=1}^{l} ( \| \mu_i(\hat{x})-\hat\mu_i \|_2 + \| \sigma_i^2(\hat{x})-\hat\sigma_i^2\|_2 )
\end{equation}
where $\mu_i(\hat{x})$ and $\sigma_i^2(\hat{x})$ are the mean and variance of output feature maps from each network layer $i$, and $\hat\mu_i$ and $\hat\sigma_i^2$ denote the running mean and variance of pre-trained network.

\paragraph{Image Prior Regularizer.} DeepDream\cite{mordvintsev2015deepdream} uses image prior regularizers
 forcing synthetic images to be stably optimized following pre-defined prior:
\begin{equation}\label{eq4}
\mathcal{R}_{prior} = \lambda_{TV} \mathcal{R_{TV}}(\hat{x}) + \lambda_{l_{2}}\mathcal{R}_{l_{2}}(\hat{x})
\end{equation}
$\mathcal{R_{TV}}$ penalizes the sparsity in synthesized images with scaling factor $\lambda_{TV}$. $l_2$ normalization, $\mathcal{R}_{l_{2}}(\hat{x})=\|\hat{x}\|_{2}$, encourages the synthesized images to have a small norm with scaling factor $\lambda_{l_{2}}$.

\subsection{Inversion Using Generative Model}\label{sec3.2}
Our goal is to encourage the generator to implicitly approximate the original dataset distribution $p_{real}(x)$.
We use the conventional conditional GAN \cite{mirza2014conditional} concept: the objective of the generator is to map $z$ from $p_z(z)$ to original data space $\mathcal{X}$ from $p_{real}(x|y)$. We sample the latent vector $z$ from the normal distribution $\mathcal{N}(0, 1)$ concatenated with $y$ encoded as one-hot vector.
However, if the latent vector changes every training epoch, the latent vectors are largely ignored or have minor effects on the variations of the images in our framework.
To address the mode-collapse problem, we propose a ``one-to-one" approach, which utilizes one generator for synthesizing one batch of samples during the current batch synthesis step $B$.
Then, after sufficiently optimizing $\theta_G$ for a latent vector $z$, we re-sample $z'$ and re-initialize the generator weight from $\theta_G$ to $\theta'_G$. As a result, the continuously re-initialized generator specifically captures a unique sample $x$ corresponding to each $z$ by optimizing the $\theta_G$. Therefore, our generator can synthesize images from various modes with respect to the different $B$, leading to diverse images. The generator architecture are shown in the appendix.

\subsection{Feature Transfer Pyramid}\label{sec3.3}
Our goal is to utilize the richer image prior from the pre-trained classifier. Thus,
we propose \textit{Feature Transfer Pyramid}(FTP) as a sub-generator, a novel hierarchical schema to gradually capture the distribution of the original dataset from different scale feature maps. As shown in Fig.\ref{Figure1}, FTP operates in conjunction with a pre-trained classifier. More specifically, given the generator output $G(z|y)$, we feed it into the pre-trained classifier and extract the feature maps $F$=$\{f_1, f_2, f_3,...f_N\}$ with different layers. The $f_1$ denotes the low-level feature maps, and $f_N$ is higher-level feature map. We extract the feature maps at the downsampling point because different scale feature maps have different representations of a pre-trained classifier. The low-level feature maps include the simple and primary characteristics, and the higher-level feature map increasingly represents complex semantic representation.
Fig.\ref{Figure1}(c) shows the building block of FTP. Now, we can get the outputs of each block, the Feature enhancement map $M$=$\{m_1, m_2,...m_L\}$, as:
\begin{equation}\label{FTPoutput}
    m_l= 
\begin{cases}
    W_{1}^{l}(\Phi(f_{N})),& \text{if } l = 1\\
    W_{1}^{l}(\Phi(W_{3}^{l}(m_{l-1} \oplus f_{N-l}))),              & \text{otherwise}
\end{cases}
\end{equation}
where $\Phi(\cdot)$ denotes the upsample layer, $W_{1}^l$ is the $l_{th}$ $1\times1$ convolution layer, and $W_{3}^l$ denotes the $l_{th}$ $3\times3$ convolution layer of FTP. First, $f_N$, the last feature map, is upsampled to fit the first FTP output size and fed to FTP as input. Then, the output of FTP block is summed with the $f_{N-1}$ which is a prior stage feature map from a classifier. 
Each output of FTP block is gradually added to feature maps at the corresponding stage of the classifier and undergoes convolution operations except the first block. Finally, the characteristic of original data is enhanced in the synthesized images by adding the $m_L$ to $G(z|y)$ element-wisely. FTP gradually captures the different spatial characteristics in the target class, and it leads to synthesized images which resembles the real data. We build up the layers of FTP when the spatial resolution change.
More detailed structure of FTP are in the appendix.
\begin{figure}[!t]
\centering
    \includegraphics[width=0.85\columnwidth]{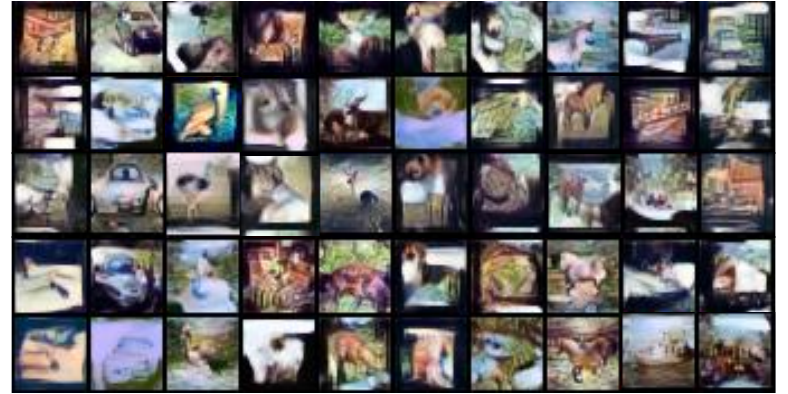}
    \hfil
\caption{One random batch samples of CIFAR-10 images. Our proposed method can synthesize the diverse images within one batch. Each column represents the same target class image. The objects in mini-batch images has different shapes and colors.}
\label{Figure3}
\end{figure}

\subsection{Adaptive Channel Scaling}
The pre-trained classifier is trained with dataset which is scaled to [0,1] and normalized with its mean and variation. Thus, the pre-trained classifier is biased to channel scale range of normalized original data, which can be used as image prior. To implicitly utilize the optimal scale range of original data, we propose the \textit{Adaptive Channel Scaling} parameters $\alpha$, which are learnable.
\begin{equation}\label{eq6}
\alpha \in \mathbb{R}^{B \times C \times 1 \times 1}
\end{equation}
Where the $B$ denotes the batch size, and the value of $C$ is 3 known as RGB channel. We iteratively multiply $\alpha$ to final images during inversion steps to adjust the output channel distribution of synthesized images. Finally, we can get the final synthesized images $\hat{I}$ using below equation:
\begin{equation}\label{eq7}
\hat{I_i} = \alpha_i \otimes (m_L \oplus G(z|y))
\end{equation}
where $\oplus$ and $\otimes$ means element-wise add/multiply individually, and $m_L$ is the last Feature enhancement map in Eq.~\ref{FTPoutput}.
We multiply $\alpha$ to combination map with generator output and $m_L$. This ACS parameters induce the synthesized images to the ``optimal scale range" which leads to produce the suitable loss for synthesizing original distribution.
Finally, we again feed these final images to the pre-trained classifier $T$ and produce the gradient. Our generator, FTP and $\alpha$ are optimized by same objective function which is defined identically as Eq.~\ref{objective}.
\begin{equation}\label{eq5:Equation8}
\mathcal{L}_{inv}(\hat{I},y)=\mathcal{L}_{CE}(\hat{I},y)+\mathcal{R}_{BN}(\hat{I})+\mathcal{R}_{prior}(\hat{I})
\end{equation}
After finishing the inversion epochs, the generator, FTP and $\alpha$ are re-initialized and synthesize the next batch of images. The overall process of NaturalInversion is summarized in Alg.~\ref{inversion_alg}.

\begin{algorithm}[t]
\newcommand{\factorial}{\ensuremath{\mbox{\sc Factorial}}}
	\caption{NaturalInversion Algorithm}\label{inversion_alg}
	\begin{algorithmic}[1]
	\Statex \textbf{Require:} A pre-trained teacher network $T$
	\Statex \textbf{Output:} Inversion Images $\hat{I}$
		\For {number of batches $B$}
		    \State {initialize Generator $G(\cdot ; \theta_{G})$, FTP $P(\cdot;\theta_{P})$, $\alpha$}
		    \State {$z \leftarrow$ $\mathcal{N}(0,1)$}
		    \For {inversion epoch $E$}
		        \State {$\hat{x} \leftarrow$ $G(z|y)$} \Comment{generator output}
		        \State {$m_L \leftarrow$ $P(\hat{x})$} \Comment{FTP output}
		        \State {$\hat{x}' \leftarrow$ $m_L + \hat{x}$}\Comment{add last FTP output to $\hat{x}$}
		        \State {$\hat{I} \leftarrow$ $\alpha \times \hat{x}'$} \Comment{synthesize the final image}
		        \State {$\mathcal{L}_{inv} \leftarrow$ $T(\hat{I})$ by Eq.\ref{eq5:Equation8}}  \Comment{compute loss function}
		        \State {$\theta_{G} \leftarrow$ $\theta_{G}-\eta_{G} \nabla{\theta_{G}}\mathcal{L}_{inv}$}
		        \State {$\theta_{P} \leftarrow$ $\theta_{P}-\eta_{P} \nabla{\theta_{P}}\mathcal{L}_{inv}$}
		        \State {$\alpha \leftarrow$ $\alpha-\eta_{\alpha} \nabla{\alpha}\mathcal{L}_{inv}$}
		    \EndFor
		    \State \Return{Mini batch Images $\hat{I}$}
		\EndFor
	\end{algorithmic} 
\end{algorithm}

\section{Experiments}
In this section, we evaluate the performance of NaturalInversion on CIFAR-10 and CIFAR-100 \cite{Krizhevsky_2009_17719}. Our experiments contain two parts, (1) analysis of our method: we verify that our method can synthesize more natural images (2) applications: we ensure the effectiveness of NaturalInversion on various applications in data-free conditions. Our experiments settings can be found in appendix.

\subsection{Analysis of NaturalInversion}\label{sec4.1}
We perform several studies to verify that synthesized images using our methods capture the original dataset distribution. As part of our experiments, we compare the (a) visualization, (b) t-SNE, (c) comparison of generative model evaluation metric result and (d) from-scratch training experiments.

\paragraph{Visualization.}
We show that our synthesized images of CIFAR-10 are more visually realistic than prior works in Fig.~\ref{Figure2}.
This is because FTP further captures the characteristics of the target class: the shape or color of individual samples are more clearly reflected in synthesized images. Furthermore, we show that our ``one-to-one" approach synthesizes more diverse images by optimizing particular generator for particular subset of original data as shown in Fig.~\ref{Figure3}. In conclusion, our method are more consistent with original data distribution with high fidelity and diversity than prior works.

\begin{figure}[t]
\centering
    \includegraphics[width=0.95\columnwidth]{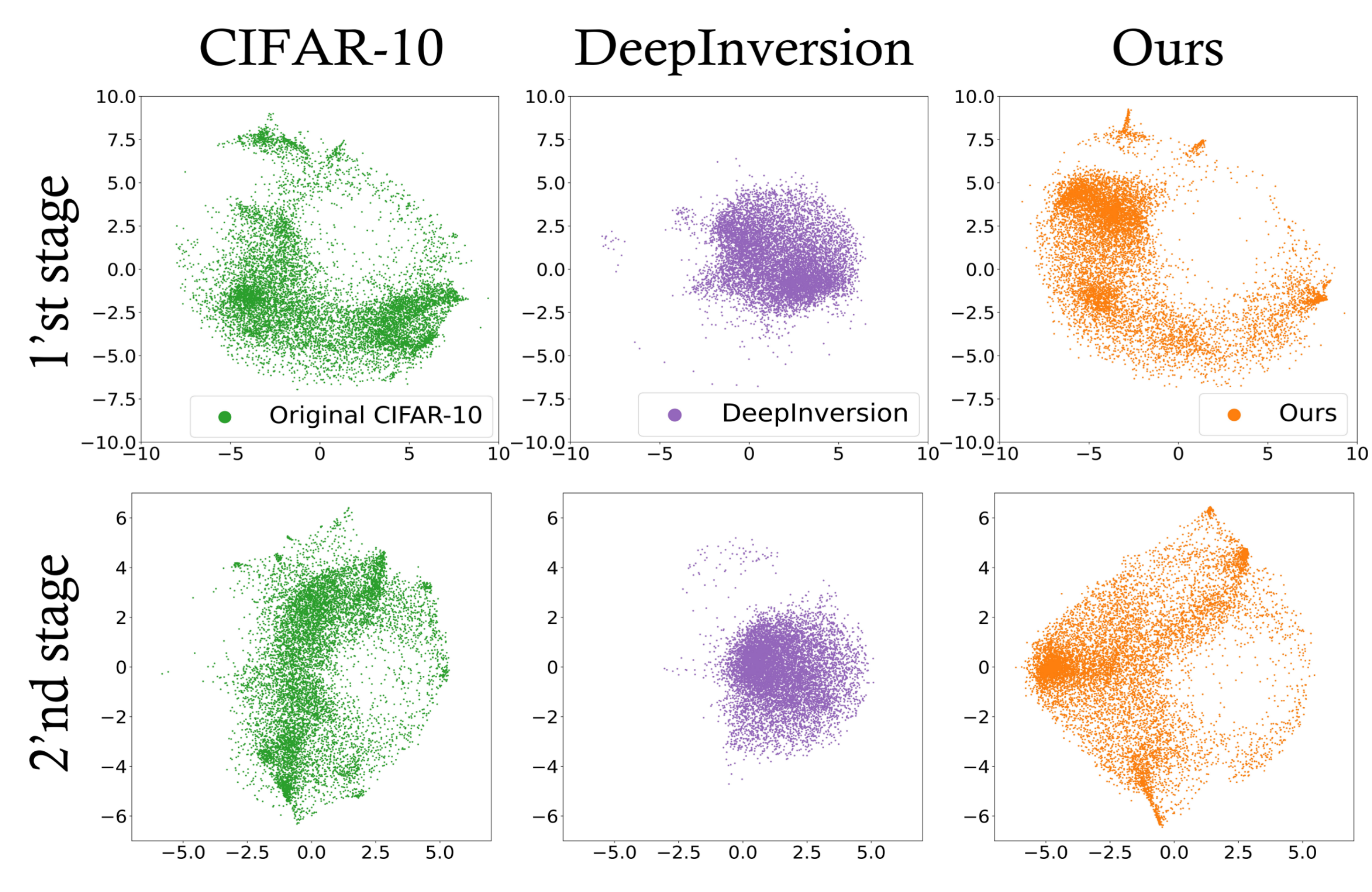}
\caption{t-SNE of original CIFAR-10, DI, and ours. We extract the feature embedding from 1st and 2nd residual blocks. The green denotes the randomly sampling 10k original CIFAR-10, The orange is NaturalInversion 10k images and DI 10k images denote purple color.}

\label{Figure4}
\end{figure}

\paragraph{Feature Visualization.}
To verify that our method encourages capturing original data distribution, we visualize the feature space from each residual block output and embedding layer of ResNet-34 using t-SNE \cite{van2008visualizing}. For visualizing feature space, we use the ResNet-34 trained by CIFAR-10 and compare the low-level feature space among 10k original CIFAR-10, DI, and our method. Entire t-SNE results are in the appendix. As shown in Fig.~\ref{Figure4}, 
the feature representations of DI have low diversity and are heavily different from the original distribution.
In contrast, the feature representation of our method is more similar to original data than DI, which verifies that our method fairly estimates the internal feature representation of real data. 

\begin{table}[t]
\centering
\resizebox{.45\textwidth}{!} {%
    \begin{tabular}{clcccc}
    
    \noalign{\global\arrayrulewidth0.03cm}
    \hline
    \noalign{\global\arrayrulewidth0.4pt}
    
    Dataset                         & Method  & IS      & FID       & Precision & Recall    \\ \hline
    
    \multirow{3}{*}{CIFAR-10}       & WGAN-GP & 7.86    & 29.3      & 0.7040    & 0.4353    \\ \cline{2-6}
                                    & DI      & 2.67    & 197.33    & 0.5996    & 0.0028    \\ 
                                    & Ours    & 5.15    & 76.04     & 0.6797    & 0.2792    \\ \hline
        
    \multirow{3}{*}{CIFAR-100}      & WGAN-GP & 6.68    & 37.30     & 0.7007    & 0.4023    \\ \cline{2-6}
                                    & DI      & 4.04    & 151.52    & 0.4051    & 0.0119    \\ 
                                    & Ours    & 6.40    & 62.90     & 0.6845    & 0.2894    \\ \hline
                                    
    \end{tabular}
}
\caption{Metric result of synthesized images by WGAN-GP, DI, and ours. A higher score of IS, Precision and Recall is better whereas a lower score of FID is better.}
\label{Table1}
\end{table} 

\begin{table}[t]
    \centering
    \begin{subtable}
    \centering
    \resizebox{.45\textwidth}{!} {%
    \begin{tabular}{clccc}
    \noalign{\global\arrayrulewidth0.03cm}
    \hline
    \noalign{\global\arrayrulewidth0.4pt} 
    
    Inversion Model                 & Train Model   & DAFL & DI     & Ours \\
    
    \noalign{\global\arrayrulewidth0.03cm}
    \hline
    \noalign{\global\arrayrulewidth0.4pt} 
    
    \multirow{3}{*}{ResNet-34}       & ResNet-18    & 32.40 & 49.61 & \textbf{74.55} \\
                                     & VGG-11       & 15.85 & 39.46 & \textbf{67.31} \\ 
                                     & MobileNetV2  & 33.49 & 34.86 & \textbf{64.34} \\ \hline
                                     
    \multirow{3}{*}{VGG-16}          & ResNet-18    & 1.96  & 46.90 & \textbf{72.38} \\
                                     & VGG-11       & 1.90  & 41.94 & \textbf{67.95} \\
                                     & MobileNetV2  & 2.11  & 31.32 & \textbf{62.86} \\ \hline
    \end{tabular}
    }
    \end{subtable}%
\caption{From scratch experiments of CIFAR-100. We train the various models from scratch using synthesized images without any information about the training set or assistance from other pre-trained networks.}
\label{Table2}
\end{table}


\begin{table}[t]
\centering
\resizebox{.47\textwidth}{!} {%
\begin{tabular}{c@{\quad}c@{\quad}c|rrrrr}
\noalign{\global\arrayrulewidth0.03cm}
\hline
\noalign{\global\arrayrulewidth0.1pt}

O2O & FTP & ACS & K.D acc & IS & FID & Precision & Recall \\

\noalign{\global\arrayrulewidth0.03cm}
\hline
\noalign{\global\arrayrulewidth0.5pt}

\multicolumn{3}{c|}{DeepInversion}  & 42.22\%    & 4.04 & 151.52 & 0.4051 & 0.0119 \\ \hline
\noalign{\global\arrayrulewidth0.05pt}
           &            &               & 19.96\%   & 4.29 & 87.09 & 0.7483 & 0.0118 \\
\checkmark &            &               & 56.69\%   & 5.51 & 65.12 & 0.7732 & 0.1529 \\
\checkmark & \checkmark &               & 58.61\%   & 5.41 & \textbf{63.36} & \textbf{0.7829} & 0.1617 \\
\checkmark &            & \checkmark    & 59.42\%   & 6.31 & 65.34 & 0.6589 & 0.2845 \\ \hline
\checkmark & \checkmark & \checkmark    & \textbf{60.57\%} & \textbf{6.44} & 63.54 & 0.6692 & \textbf{0.2960} \\

\hline

\end{tabular}}
\caption{Effectiveness of DeepInversion and each component of our method. Our method outperforms DI in IS/FID scores and K.D with a small number of images (50k)}
\label{Table5}
\end{table}

\begin{table}[ht]
\centering
\resizebox{0.44\textwidth}{!} {%
\begin{tabular}{@{}ccccc@{}}
\midrule
Feature Stage & $f_4$  & $f_3$,$f_4$ & $f_2$,$f_3$,$f_4$    & use all (ours)       \\ \midrule
1'st stage    & 0.6892 & 0.6845      & 0.6736               & 0.6634                \\
final stage   & 77.90  & 75.84       & 75.52                & 73.72                 \\ \midrule
\end{tabular}}
\caption{FID score under the different usage of feature maps. We gradually choose the feature maps of ResNet-34 from high-level($f_4$) to low-level feature maps($f_1$).}
\label{Table3}
\end{table}

\paragraph{Generative Model Evaluation Metric.} 
We further analyze our methods with the generative model evaluation metric to assess how similar our images are to the original dataset : (a) single-value metric - Inception Score (IS) \cite{salimans2016improved} and Frechet Inception Distance (FID) \cite{heusel2017gans}. (b) two-value metrics- Precision and Recall (P\&R) \cite{sajjadi2018assessing}. We specifically verify the qualitative fidelity and diversity of images by the two-value metrics. We synthesize 50k CIFAR-10/100 images from ResNet-34 and use the logit and embedding layer of ImageNet pre-trained Inception-v3 \cite{szegedy2016rethinking} for calculating the IS and remaining metrics respectively. To show how well we approximated the original data distribution, we compare the synthesized images with the images by DI and GAN-based model, WGAN-GP, which utilizes the original data.
As shown in Table \ref{Table1}, NaturalInversion outperforms DI across CIFAR-10/100 in terms of IS, FID, P\&R. In addition, NaturalInversion is even close to WGAN-GP baselines without original data. Through these evaluation metrics, we confirm that our method estimates the original distribution well without the real data. 

\begin{table*}[!t] 
\centering
\resizebox{0.9\textwidth}{!} {%
\begin{tabular}{cccrrrrrrrr}
\noalign{\global\arrayrulewidth0.04cm}
\hline
\noalign{\global\arrayrulewidth0.1pt}

\multirow{2}*{\textbf{Dataset}} &\multirow{2}*{\textbf{Teacher}}  & \multirow{2}*{\textbf{Student}}    & \multicolumn{8}{c}{\textbf{Accuracy}} \\
\noalign{\global\arrayrulewidth0.01cm}
\cline{4-11}
\noalign{\global\arrayrulewidth0.1pt}
                            &&& \textbf{T.}                      & \textbf{S.}    
                            & \textbf{DAFL}                      & \textbf{DFAD}           
                            & \textbf{DI}                        & \textbf{ADI}                           
                            & \textbf{ours}                      & \textbf{A-ours}   \\
        
\noalign{\global\arrayrulewidth0.03cm}
\hline
\noalign{\global\arrayrulewidth0.4pt}

\multirow{6}*{CIFAR-10}
&ResNet-34 & ResNet-18      & 95.57                     & 95.20 
                            & 92.22                     & 93.30
                            & 91.43                     & 93.26       
                            & \underline{93.72}         & \textbf{94.87}        \\
                        
&ResNet-34 & VGG-11         & 95.57                     & 92.44 
                            & $_{}^{*}\textrm{44.30}$   & $_{}^{*}\textrm{\underline{88.66}}$
                            & $_{}^{*}\textrm{83.49}$   & $_{}^{*}\textrm{75.53}$
                            & \textbf{89.41}            & 88.46                 \\
                        
&ResNet-34 & MobileNetV2    & 95.57                     & 94.69 
                            & $_{}^{*}\textrm{72.16}$   & $_{}^{*}\textrm{92.71}$     
                            & $_{}^{*}\textrm{91.00}$   & $_{}^{*}\textrm{93.64}$       
                            & \underline{92.96}         & \textbf{94.06}        \\

&VGG-11   & ResNet-18       & 92.44                     & 95.20 
                            & $_{}^{*}\textrm{84.19}$   & $_{}^{*}\textrm{89.35}$         
                            & 83.82                     & \underline{90.36}
                            & 90.10                     & \textbf{91.84}        \\
                            
&VGG-11   & VGG-11          & 92.44                     & 92.44 
                            & $_{}^{*}\textrm{82.18}$   &$\textbf{$_{}^{*}\textrm{91.34}$}$
                            & 84.16                     & 90.78
                            & 89.79                     & \underline{91.07}     \\
                        
&VGG-11   & MobileNetV2     & 92.44                     & 94.69 
                            & $_{}^{*}\textrm{54.79}$   & $_{}^{*}\textrm{84.96}$         
                            & $_{}^{*}\textrm{87.77}$   & $_{}^{*}\textrm{88.97}$       
                            & \underline{89.59}         & \textbf{90.62}        \\ \hline

\multirow{6}*{CIFAR-100}
&ResNet-34 & ResNet-18      & 78.02                     & 76.87 
                            & \textbf{74.47}            & 67.70
                            & $_{}^{*}\textrm{45.91}$   & $_{}^{*}\textrm{64.38}$       
                            & 67.00                     & \underline{72.82}      \\

&ResNet-34 & VGG-11         & 78.02                     & 68.64 
                            & $_{}^{*}\textrm{48.43}$   & $_{}^{*}\textrm{20.61}$        
                            & $_{}^{*}\textrm{36.04}$   & $_{}^{*}\textrm{51.06}$       
                            & \underline{61.96}         & \textbf{65.37}        \\

&ResNet-34 & MobileNetV2    & 78.02                     & 68.02
                            & $_{}^{*}\textrm{59.46}$   & $_{}^{*}\textrm{57.30}$
                            & $_{}^{*}\textrm{44.30}$   & $_{}^{*}\textrm{58.88}$
                            & \underline{66.42}          & \textbf{71.26}        \\
                            
&VGG-16    & ResNet-18      & 73.75                     & 76.87 
                            & $_{}^{*}\textrm{24.91}$   & $_{}^{*}\textrm{53.41}$
                            & $_{}^{*}\textrm{50.32}$   & $_{}^{*}\textrm{56.36}$
                            & \underline{66.32}         & \textbf{69.84}         \\ 

&VGG-16    & VGG-11         & 73.75                     & 68.64 
                            & $_{}^{*}\textrm{23.96}$   & $_{}^{*}\textrm{44.34}$
                            & $_{}^{*}\textrm{46.24}$   & $_{}^{*}\textrm{45.14}$
                            & \underline{61.68}         & \textbf{64.37}         \\ 

&VGG-16    & MobileNetV2    & 73.75                     & 68.02
                            & $_{}^{*}\textrm{9.00}$    & $_{}^{*}\textrm{41.61}$        
                            & $_{}^{*}\textrm{37.76}$   & $_{}^{*}\textrm{54.38}$       
                            & \underline{64.04}         & \textbf{66.95}         \\ \hline

%

\end{tabular}}
\caption{The results of data-free knowledge distillation on CIFAR-10/100 with synthesized images from various inversion methods. 
The bold type is the best value of the same architecture experiment, and the underbar type is the second-best value. T. and S. means the baseline accuracy of teachers and students trained on the original training set. *: our re-implementations.}
\label{KDtable}
\end{table*}

\paragraph{From Scratch Training.}
To demonstrate how well our method captures the original distribution, we train the model from scratch with images by our method without any real data and compare the training accuracy of model with other methods: DAFL and DI.
First, we synthesize 256k images using DAFL, DI, and our methods from ResNet-34, VGG-16 trained by CIFAR-100 dataset. Then, we train the randomly initialized networks from scratch using synthesized images. We set the mini-batch size as 128 and train the model for 200 epochs using an SGD optimizer with a 0.05 learning rate. Finally, we evaluate the training accuracy by forwarding the CIFAR-10/100 training set to trained classifiers. Table \ref{Table2} depicts how the images from NaturalInversion restores the accuracy of model compared to other methods, meaning that our method successfully captures the original training set distribution.
Our method outperforms prior works with the training accuracy by a large gap. 
In addition, we observed that our approach ensures high accuracy, even though the types of training models are different from the inversion model, implying that our method produces more ``generalized" images that approximate the original data distribution. The test accuracy result of CIFAR-10/100 is available in the appendix.

\subsection{Ablation Study}
\paragraph{Effectiveness of Each Component in NaturalInversion.}
We conduct the ablation experiments to understand how each component affects our overall method.
We synthesize 50k images from ResNet-34 trained by CIFAR-100 for each configuration. We perform the knowledge distillation to the ResNet-18 student network, and evaluate the quality of synthesized images using FID, IS, and P\&R. All settings in image synthesis are the same as Section 4.1. Table \ref{Table5} reports the result of ablation experiments. 
The \textit{one-to-one} generator can reduce the mode collapse of synthesized images. Also, ACS parameters per each instance image affect the image's color. These components increase the student network accuracy by improved diversity. FTP leads to synthesizing higher fidelity images by using the feature maps, achieving the best FID and Precision. 
In summary, FTP improves fidelity, and one-to-one generator and ACS help diversity. The accuracy of the student network achieves the best accuracy, 60.57\%, when using all components. 
\begin{figure}[!t]
\includegraphics[width=0.95\columnwidth]{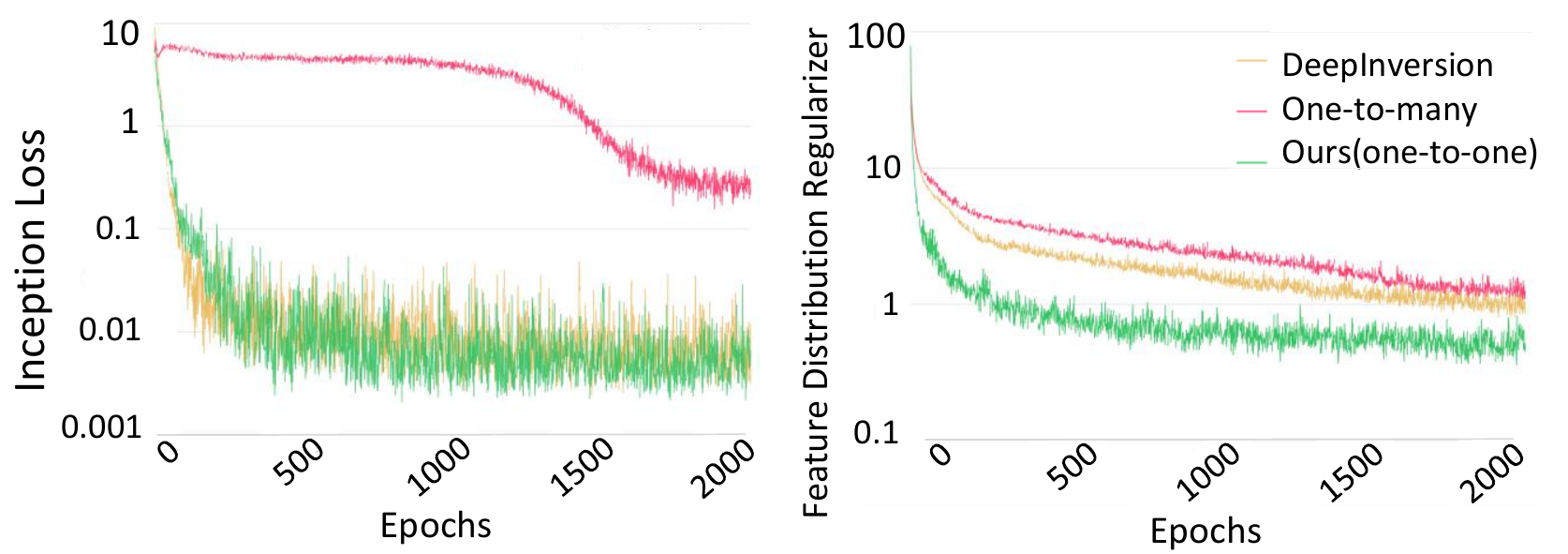}
    \caption{Training loss curve during CIFAR-10 image synthesis. Inception loss (left) and feature distribution regularizer (right). One-to-Many denotes that non-linearity optimization without one-to-one approach.}
    \label{Figure5}
\end{figure}
\paragraph{Bottom-Up Usage of Feature Map.}
To verify the effectiveness of our \textit{Feature Transfer Pyramid}, we gradually increase feature maps combined to FTP. We observe how close the distribution of synthesized images is to the original distribution according to the number of feature maps. We continuously select high-level and low-level feature maps, then build up the FTP blocks. First, we synthesize 10k images from the ResNet-34 trained CIFAR-10 on each case. Then, we calculate FID scores using the embedding layer of the Inception-v3. As shown in Table \ref{Table3}, we demonstrate that the more we gradually stack FTP block, the lower the FID score. Since each block of FTP enhances the corresponding characteristics of the original target class, we synthesize the images closer to the original data.

\paragraph{Training Loss Curve.}
To test that our generative model with \textit{one-to-one} approach helps to converge to a specific optimal point faster, we test training loss convergence of three cases: (a) optimization without non-linearity (b) non-linear optimization without \textit{one-to-one} approach (c) our method. Fig.\ref{Figure5} shows the training loss curve of 
$\mathcal{L}_{CE}$ and 
$\mathcal{R}_{BN}$ that play a key role in generating images in NaturalInversion. We simultaneously plot the training loss curve during 2k CIFAR-10 image inversion epoch. Because of the difficulty of optimizing the input space without non-linearity, DI converges slower than our method, and the final training loss is greater than ours for both $\mathcal{L}_{CE}$ and $\mathcal{R}_{BN}$. Moreover, using the generator to induce non-linearity helps the convergence by greater ability to represent the complex characteristics of the original distribution. 
In the end, we maximize the improvement in converge speed and training loss when using \textit{one-to-one} generator, leading to eased optimizing process.

\subsection{Application 1: Knowledge Distillation (KD)}
In this section, we ensure that we can transfer information from a teacher network to a student network without original data and outperform existing data-free knowledge distillation approaches. We compare our approach with different methods: DAFL, DFAD, DI, and ADI \cite{yin2020dreaming}. Especially, ADI improves upon DI with increment in diversity by the teacher-student disagreement.
Since our method can be applied to existing frameworks of inversion, we combine our approach with the ADI method to improve the knowledge distillation performance. For CIFAR-10 KD, we use ResNet-34, VGG-11 as a teacher network, and use ResNet-34 and VGG-16 for CIFAR-100 KD.  
To verify our method synthesizes less biased images to the teacher, which means generalized images regardless of the model type, we distill knowledge to various students from each teacher. We choose the ResNet-18 \cite{he2016deep}, VGG-11, and MobileNet-V2 \cite{sandler2018mobilenetv2} as a student network. 
We synthesize the 256k images from the teacher network and use all images to train the student network. 
The results for the data-free knowledge distillation are reported in Tab.~\ref{KDtable}, showing that our approach achieves comparable performance compared to the most of other methods.  
Furthermore, although the teacher and student network have different architecture types, NaturalInversion still achieves high performance over other methods, indicating that our method produces less model-biased images regardless of the data and network type. 

\begin{figure}[t]
        \subfigure{
            \includegraphics[height=0.32\columnwidth]{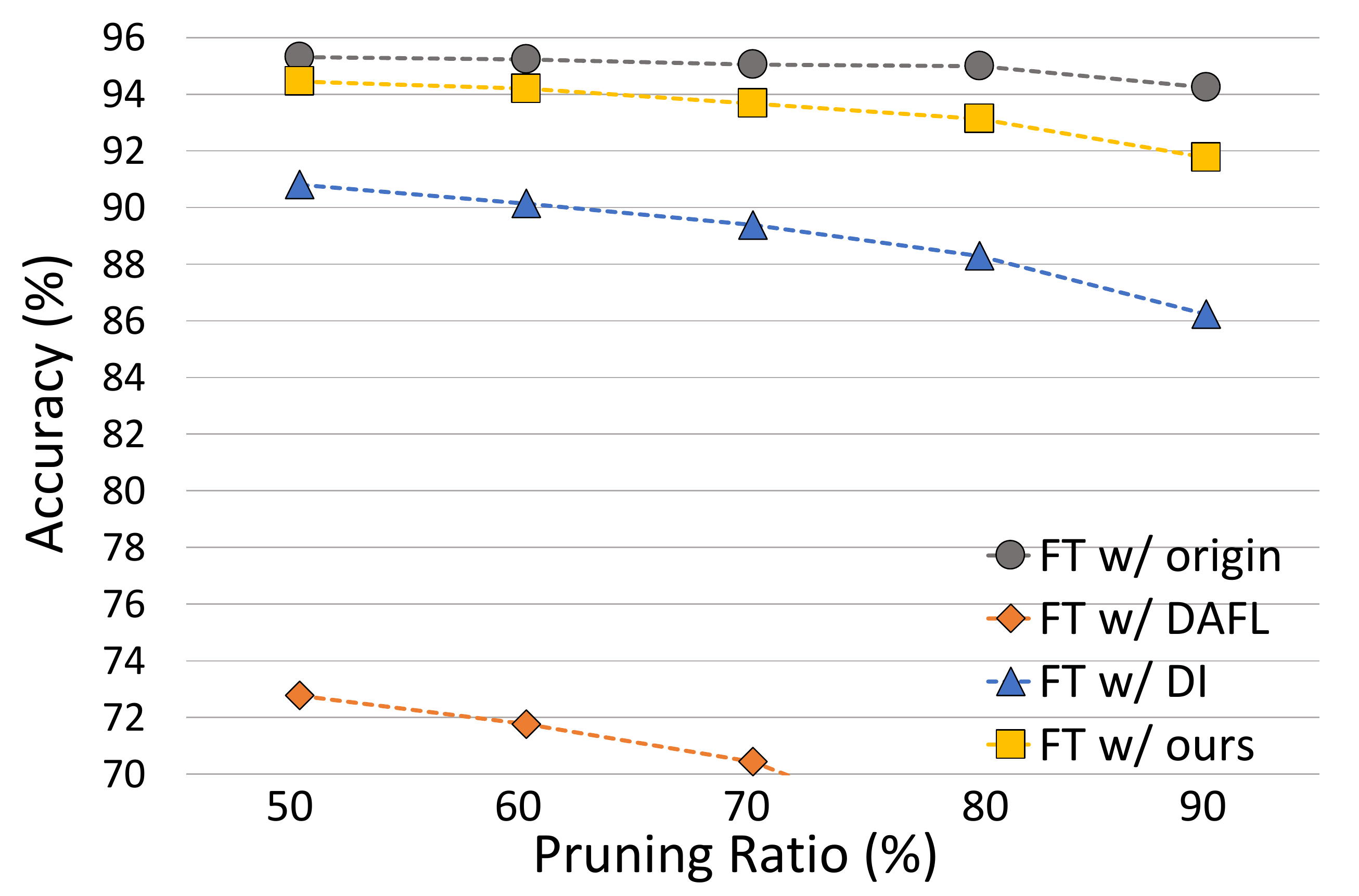}}
        \subfigure{
            \includegraphics[height=0.32\columnwidth]{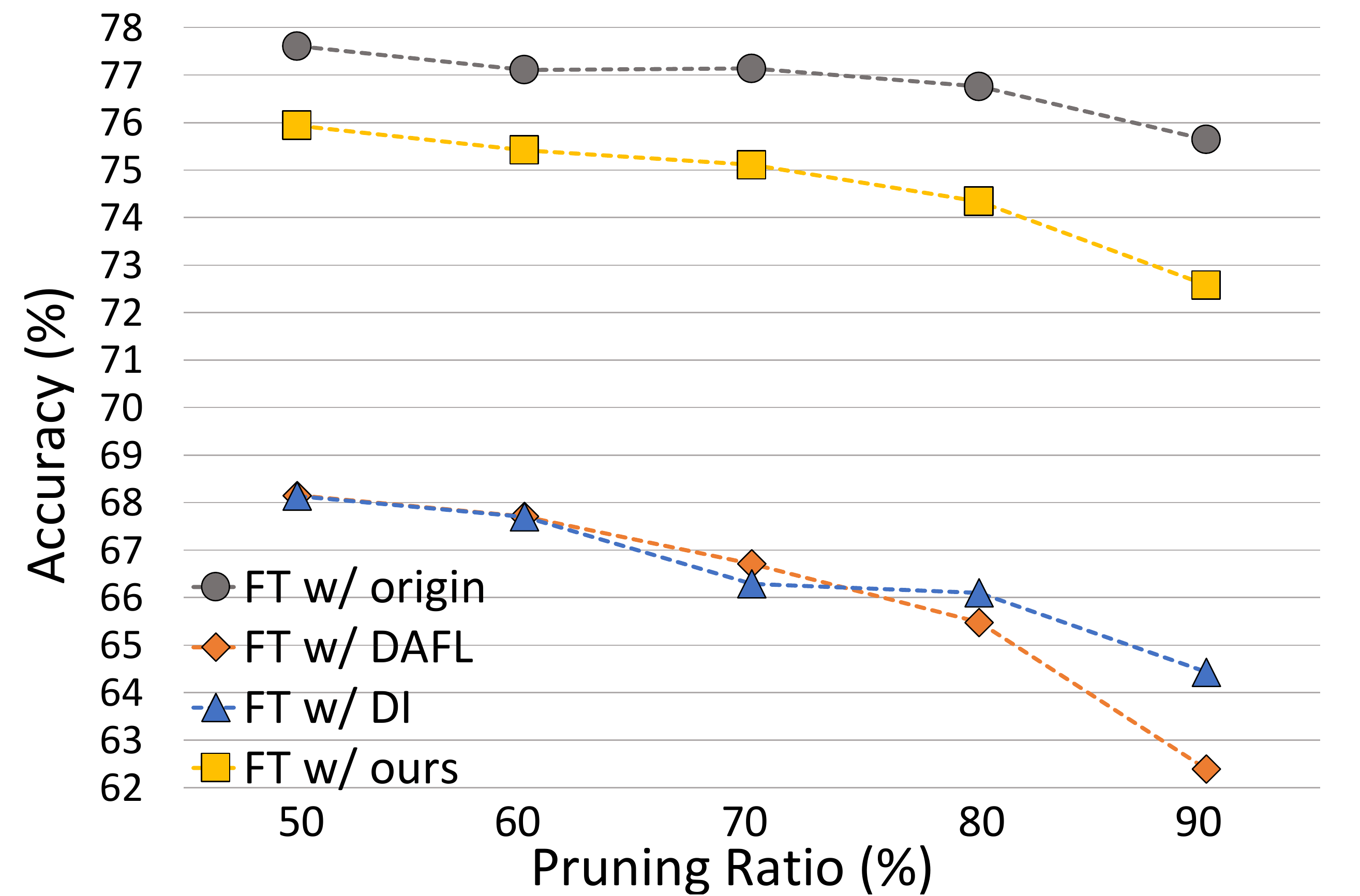}}
    \caption{Pruning performance of ResNet-34 on CIFAR-10 (left) and CIFAR-100 (right) under different pruning ratios. We synthesize images from ResNet-34 and fine-tune the pruned model using synthesized images by various methods.}
    \label{Figure6}
\end{figure}

\subsection{Application 2 : Pruning} 
We investigate that our method can improve the pruned model accuracy without real data. For the pruning, we utilized L1-norm pruning criteria that prunes a certain percentage of filters with smaller L1-norm \cite{liu2017learning}. We carry out experiments on different pruning ratios from 50\% to 90\%, and all experiments are performed with ResNet-34. Our pruning setup is the same with \citet{liu2018rethinking}, and we locally prune the least important channels in each layer by the same pruning ratio. After pruning ResNet-34 on CIFAR-10/100, we fine-tune the pruned model using synthesized images from the baseline model for 20 epochs by SGD with 0.001 learning rate. As shown in Fig.\ref{Figure6}, DAFL has poor performance with 58.34\% at a pruning ratio 90\% on CIFAR-10 because it synthesizes images for a specific purpose, knowledge distillation. DI has low performance on CIFAR-100 because the more complexity the image, the more difficult it is to optimize without non-linear characteristics. In contrast, we achieve the best accuracy recovery of a pruned model up to 11.21\% improvement on 90\% pruning ratio due to the high fidelity and diversity images. This result ensure that our approach reflects the statistics of original data.

\section{Conclusion}
In this paper, we improve the quality of the synthesized image compared to conventional inversion methods with our proposed approach: NaturalInversion. 
First, we enhanced the characteristics of target class via \textit{Feature Transfer Pyramid} by using multi-scale feature maps from classifier. Second, we used \textit{one-to-one} generator for alleviating the mode collapse problem and bring the non-linearity. Lastly, we proposed \textit{Adaptive Channel Scaling} parameters to implicitly learn the optimal channel scale range which has been learned by the classifier.
Through extensive experiments, we demonstrated the effectiveness of NaturalInversion. Our methods not only capture the original data distribution but also are generalized, less biased to inversion model. 
We hope this work helps for the further progress in synthesizing realistic images in data-free conditions. 


\section*{Acknowledgements}
This work was supported by the National Research Council of Science \& Technology (NST) grant by the Korea government (MSIT) [CRC-20-02-KIST], and by Institute of Information \& communications Technology Planning \& Evaluation (IITP) grant funded by the Korea government (MSIT) (No.2021-0-00456, Development of Ultra-high Speech Quality Technology for Remote Multi-speaker Conference System). 

\bibliography{aaai22.bib}
\clearpage
\appendix
\section{Appendix}

\subsection{Generator Architecture for CIFAR-10/100}
We use the same architecture on CIFAR-10/100 image generation, and generator architecture can be shown in Table.\ref{generator}. We use the latent space $z$ from conditional distribution given arbitrary label $\hat{y}$, so the $C$ means the number of classes. (For CIFAR-10 image synthesis, $C$=10, and $C$=100 for CIFAR-100). The $s$,$p$, and $c$ are the stride, padding of the convolution filter, and the number of channels. Finally, the generator synthesizes the CIFAR-10/100 images with a resolution of $32\times32$. 
\begin{table}[h]
\centering
\resizebox{.51\textwidth}{!} {
\begin{tabular}{cccll}
\cline{1-3}
\textbf{Layer}    & \textbf{Detail} & \textbf{Output Size} &  &  \\ \cline{1-3}
input             &$z \in \mathbb{R}^{1024}\sim \mathcal{N}(0, 1)$                  &$(1024+C) \times 1$ &                      &  \\ \cline{1-3}
layer1            & Linear layer                 &128 $\times$ 8 $\times$ 8                      &  &  \\ \cline{1-3}
\multirow{3}{*}{Block 1}    & BN layer (c=128) \\ 
                            & LeakyReLU (0.2)  &128 $\times$ 16 $\times$ 16 \\ 
                            & Upsample $\times$ 2     \\ \cline{1-3}
                            
\multirow{4}{*}{Block 2}    &$3\times3$ Conv (s=1,p=1),\\
                            &BN layer (c=128),  & 128 $\times$ 32 $\times$ 32 \\ 
                            & LeakyReLU (0.2),\\  
                            & Upsample $\times$ 2  \\ \cline{1-3}  
                            
\multirow{3}{*}{Block 3}    & $3\times3$ Conv (s=1,p=1), \\ 
                            & BN layer (c=64), &64 $\times$ 32 $\times$ 32 \\  
                            &  LeakyReLU (0.2)  \\\cline{1-3}
                            
{output}              &$3\times3$ Conv (s=1,p=1)   &3 $\times$ 32 $\times$ 32  \\ \cline{1-3}
\end{tabular}}
\caption{Generator architecture of NaturalInversion for CIFAR-10/100 image synthesis.}
\label{generator}
\end{table}

\subsection{Feature Transfer Pyramid Architecture}
Our \textit{Feature Transfer Pyramid} (FTP) operates in conjunction with a pre-trained classifier $T$. We usually use the ResNet-34 and VGG-11/16 as a pre-trained classifier. In these two classifiers, the downsampling point of the feature map is the same, so we set the feature extraction stage the same. Therefore, we consist of a total of four blocks of FTP, and details can be found in Table.\ref{FTP}.
 
\begin{table}[h]
\centering
\resizebox{.51\textwidth}{!} {
\begin{tabular}{cccll}
\cline{1-3}
\textbf{Layer}    & \textbf{Detail} & \textbf{$m_l$ Size} &  &  \\ \cline{1-3}
\multirow{3}{*}{Block 1}    & input $f_4$ (512 $\times$4 $\times$4), \\
                            & Upsample $\times$ 2,  &256$\times$8$\times$8 \\ 
                            & $1\times1$ Conv (512 $\rightarrow$ 256)  \\ \cline{1-3}
                            
\multirow{5}{*}{Block 2}    & input $f_3$ (256 $\times$8$\times$8), \\
                            & element-wise sum $f_3+m_1$, \\ 
                            & $3\times3$ Conv,  &128$\times$16$\times$16 \\ 
                            & Upsample $\times$ 2,\\
                            & $1\times1$ Conv (256 $\rightarrow$ 128)  \\ \cline{1-3}
                            
\multirow{5}{*}{Block 3}    & input $f_2$ (128 $\times$16$\times$16), \\
                            & element-wise sum $f_2+m_2$, \\ 
                            & $3\times3$ Conv,  &64$\times$32$\times$32 \\ 
                            & Upsample $\times$ 2,\\
                            & $1\times1$ Conv (128 $\rightarrow$ 64)  \\ \cline{1-3} 
                            
\multirow{4}{*}{Block 4}    & input $f_1$ (64 $\times$32$\times$32), \\
                            & element-wise sum $f_1+m_3$, \\ 
                            & $3\times3$ Conv,  &3$\times$32$\times$32 \\ 
                            & $1\times1$ Conv (64 $\rightarrow$ 3)  \\ \cline{1-3}

\multirow{4}{*}{output}     & input $G(z|y)$ (3 $\times$32$\times$32), \\
                            & element-wise sum $G(z|y)+m_4$, \\ 
                            & $1\times1$ Conv (3 $\rightarrow$ 3),  &3$\times$32$\times$32 \\ 
                            & Tanh activation  \\ \cline{1-3}
\end{tabular}}
\caption{Feature Transfer Pyramid architecture for CIFAR-10/100 image synthesis. Each $f_n$ means the feature map extracted from classifier and $m_l$ is each FTP block output.}
\label{FTP}
\end{table}

\subsection{Experiments Settings}
In this work, we synthesize CIFAR images from several architectures, including ResNet-34 and VGG-11/16.
The CIFAR-10/100 dataset consists of 50k training images and 10k images for validation, whose resolution is $32\times32$ and has 10 or 100 categories individually. We use the almost the same setup on CIFAR-10/100 image synthesis during inversion. Our proposed structure is composed of three components: generator, FTP, and $\alpha$ which are trained by Adam optimizer with different learning rates 0.001, 0,0005, and 0.05, respectively. The $\alpha$ is initialized on normal distribution $\mathcal{N}(5.0,1.0)$. We use the ResNet-34, VGG-11 and VGG-16 as a inversion model. We extract the feature map $f_i$ on each residual block output for ResNet and pooling layer output for VGG. 
We use $\lambda_{BN}$ = 10.0 on CIFAR-10 and $\lambda_{BN}$ = 3.0 on CIFAR-100. For both CIFAR10/100, we set $\lambda_{TV}$ = $6.0 \times 10^{-3}$ and $\lambda_{l_2}$ = $1.5 \times 10^{-5}$. The mini-batch size is 256 and we synthesize the CIFAR-10/100 images for 2k, 4k inversion epochs, respectively.

\subsection{Frechet Distance Evaluation}
We show the ``generality" of NaturalInversion by comparing Frechet distance on various models with prior works in Table.\ref{FD}. In general, Frechet Inception Distance(FID) is widely used to assess the image quality in generative models. Inception-v3 model trained on the ImageNet is commonly used as an embedding network. This study uses different type embedding networks trained on ImageNet and tests CIFAR-10/100 samples to evaluate that our method synthesizes high-quality images regardless of model type or pre-trained dataset domain. We reuse the images used in the generative model evaluation metric experiments in Section 4.1. When utilizing the model trained with ImageNet, NaturalInversion performs better with a more considerable difference than DI, and only 13.67\% difference compared to WGAN-GP (baseline) trained with original data on CIFAR-10 experiments. We also choose the embedding networks trained with the CIFAR-10, the same domain we generate, so the overall FD score is low on embedding networks trained with CIFAR-10. Since NaturalInversion and DI synthesize images from ResNet-34 trained with CIRAR-10, FD score is better than WGAN-GP when using ResNet-34 as an embedding network. We achieve 3.22\% better than WGAN-GP when using VGG-11. CIFAR-100 experiments also show similar trends in CIFAR-10 experiments. In conclusion, NaturalInversion approximates the original dataset distribution well regardless of the pre-trained classifier for inversion, and we ensure that our method synthesizes generalized images.

\begin{table}[t]
\centering
\resizebox{.47\textwidth}{!} {
\begin{tabular}{clccc}
\noalign{\global\arrayrulewidth0.03cm}
\hline
\noalign{\global\arrayrulewidth0.4pt}
Train set                   & Model     & Ours   & DI      & WGAN-GP \\ \hline

\multirow{4}{*}{ImageNet}   & VGG-11    & 370.65 & 1840.24 & 150.79  \\ 
                            & ResNet-18 & 31.26  & 250.98  & 17.59   \\  
                            & ResNet-34 & 44.86  & 375.64  & 22.28   \\ 
                            & ResNet-50 & 42.18  & 138.44  & 19.30   \\ \hline
                            
\multirow{2}{*}{CIFAR-10}   & VGG-11    & 2.53   & 12.70   & 5.75    \\ 
                            & ResNet-34 & 0.27   & 0.65    & 0.94    \\ \hline
\end{tabular}}

\centering
\resizebox{.47\textwidth}{!} {
\begin{tabular}{clccc}
\noalign{\global\arrayrulewidth0.03cm}
\hline
\noalign{\global\arrayrulewidth0.4pt}
Train set                   & Model     & Ours    & DI       & WGAN-GP \\ \hline

\multirow{4}{*}{ImageNet}   & VGG-11    & 8030.03 & 12550.93 & 3626.84 \\ 
                            & ResNet-18 & 68.84   & 91.28    & 21.12 \\  
                            & ResNet-34 & 206.28  & 293.62   & 63.27 \\ 
                            & ResNet-50 & 492.35  & 610.35   & 163.73 \\ \hline
                            
\multirow{2}{*}{CIFAR-100}  & VGG-16    & 9.64 & 18.08 & 16.81 \\ 
                            & ResNet-34 & 2.89 & 10.18 & 12.76 \\ \hline
\end{tabular}}
\caption{Frechet distance across the various backbone networks. Frechet distance result of CIFAR-10 images(top) and CIFAR-100 images(bottom)}
\label{FD}
\end{table}

\begin{figure*}[t]
\centering
    \includegraphics[width=17cm]{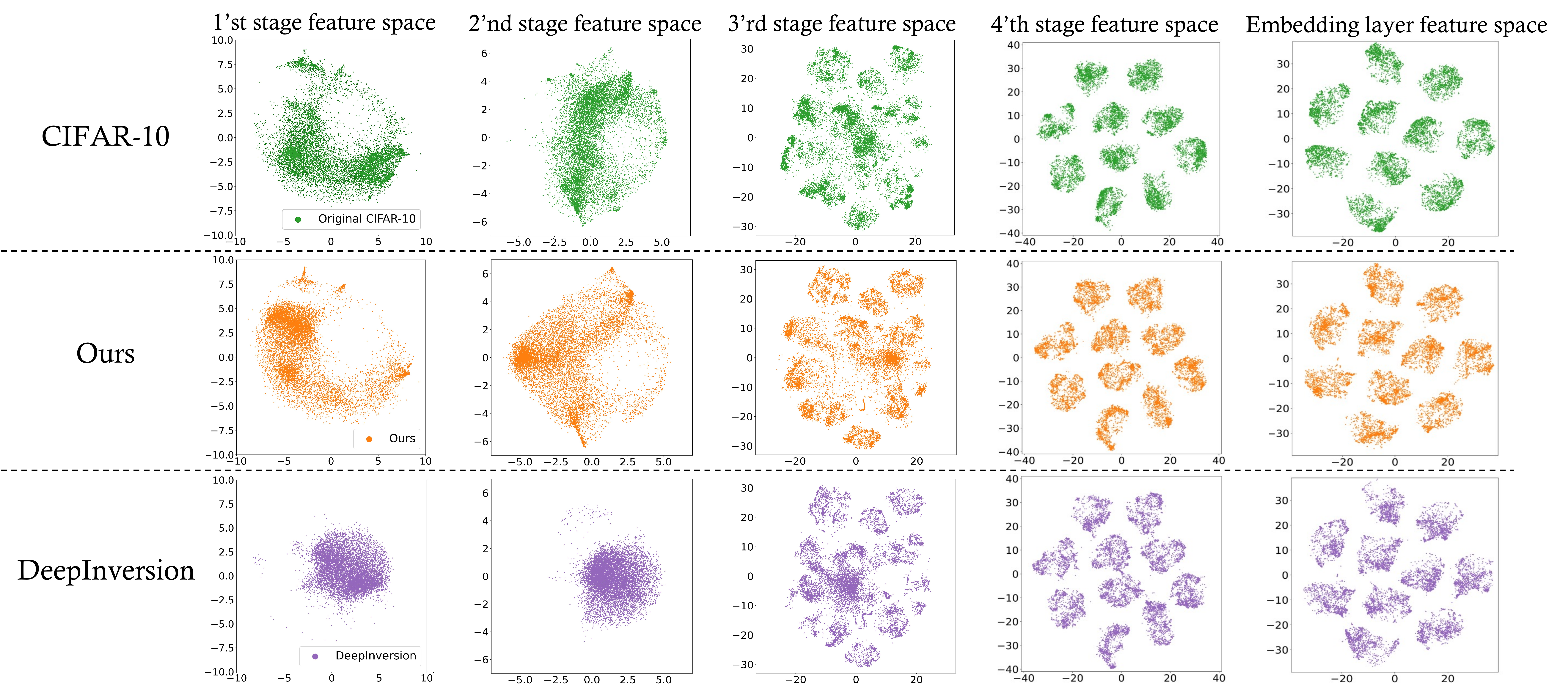}
    \hfil
\caption{t-SNE result of overall stage. We compare feature spaces with original CIFAR-10 10k and DI 10k images from low-level(1st residual block output) to high-level (embedding layer) of ResNet-34.}
\label{tSNE_all}
\end{figure*}

\begin{table*}[h]
\setlength\tabcolsep{0pt}
\setlength\extrarowheight{2pt}
\centering
\resizebox{.95\textwidth}{!} {%
\begin{tabular*}{\textwidth}{@{\extracolsep{\fill}}*{1}{clcccccc}}

\noalign{\global\arrayrulewidth0.03cm}
\hline
\noalign{\global\arrayrulewidth0.4pt} 
\setlength{\tabcolsep}{3pt}
\multirow{2}{*}{Inversion Model} & \multirow{2}{*}{Train Model} & \multicolumn{3}{c}{\textbf{Train Acc. (Loss)}}      & \multicolumn{3}{c}{\textbf{Test Acc. (Loss)}} \\ \cline{3-5} \cline{6-8}
                                 &                              & \textbf{DAFL}         & \textbf{DI}           & \textbf{Ours}         & \textbf{DAFL}         & \textbf{DI}              & \textbf{Ours}    \\
\noalign{\global\arrayrulewidth0.03cm}
\hline
\noalign{\global\arrayrulewidth0.4pt} 

\multirow{3}{*}{ResNet-34}       & ResNet-18    & 31.45(0.029) & 73.46(0.008) & \textbf{90.84}(\underline{0.003}) & 31.59(0.029) & 72.05(0.009) & \textbf{87.98}(\underline{0.004})   \\
                                 & VGG-11       & 21.04(0.027) & 67.18(0.010) & \textbf{86.76}(\underline{0.005}) & 21.52(0.027) & 65.30(0.011) & \textbf{85.74}(\underline{0.005})   \\
                                 & MobileNetV2  & 42.50(0.021) & 69.57(0.010) & \textbf{86.52}(\underline{0.004}) & 42.29(0.022) & 66.38(0.010) & \textbf{84.83}(\underline{0.005})   \\ \hline
                                 
\multirow{3}{*}{VGG-11}          & ResNet-18    & 46.98(0.018) & 80.21(0.007) & \textbf{88.13}(\underline{0.004}) & 46.08(0.018) & 78.16(0.008) & \textbf{85.53}(\underline{0.005})   \\
                                 & VGG-11       & 32.38(0.021) & 78.94(0.007) & \textbf{86.76}(\underline{0.004}) & 32.52(0.021) & 76.31(0.008) & \textbf{83.92}(\underline{0.006})   \\
                                 & MobileNetV2  & 44.17(0.017) & 73.82(0.008) & \textbf{84.03}(\underline{0.005}) & 43.52(0.017) & 72.02(0.009) & \textbf{81.40}(\underline{0.006})   \\ \hline
\end{tabular*}}

\setlength\tabcolsep{0pt}
\setlength\extrarowheight{2pt}
\centering
\resizebox{.95\textwidth}{!} {%
\begin{tabular*}{\textwidth}{@{\extracolsep{\fill}}*{1}{clcccccc}}

\noalign{\global\arrayrulewidth0.03cm}
\hline
\noalign{\global\arrayrulewidth0.4pt} 
\setlength{\tabcolsep}{3pt}
\multirow{2}{*}{Inversion Model} & \multirow{2}{*}{Train Model} & \multicolumn{3}{c}{\textbf{Train Acc. (Loss)}}      & \multicolumn{3}{c}{\textbf{Test Acc. (Loss)}} \\ 
                                                                \cline{3-5} \cline{6-8}
                                 &                              & \textbf{DAFL}& \textbf{DI}  & \textbf{Ours}         & \textbf{DAFL}& \textbf{DI}     & \textbf{Ours}  \\
                                 
\noalign{\global\arrayrulewidth0.03cm}
\hline
\noalign{\global\arrayrulewidth0.4pt} 

\multirow{3}{*}{ResNet-34}       & ResNet-18    & 32.40(0.028) & 49.61(0.019) & \textbf{74.55}(\underline{0.008}) & 30.85(0.030) & 45.18(0.021) & \textbf{65.13}(\underline{0.012})   \\
                                 & VGG-11       & 15.85(0.036) & 39.46(0.027) & \textbf{67.31}(\underline{0.012}) & 15.68(0.036) & 36.35(0.030) & \textbf{58.98}(\underline{0.017})   \\
                                 & MobileNetV2  & 33.49(0.031) & 34.86(0.025) & \textbf{64.34}(\underline{0.012}) & 30.84(0.033) & 33.30(0.027) & \textbf{59.19}(\underline{0.015})   \\ \hline
                                 
\multirow{3}{*}{VGG-11}          & ResNet-18    & 1.96(0.162)  & 46.90(0.018) & \textbf{72.38}(\underline{0.009}) & 2.08(0.163)  & 43.48(0.020) & \textbf{63.21}(\underline{0.013})   \\
                                 & VGG-11       & 1.90(0.049)  & 41.94(0.024) & \textbf{67.95}(\underline{0.012}) & 1.89(0.050)  & 38.02(0.027) & \textbf{59.26}(\underline{0.018})   \\
                                 & MobileNetV2  & 2.11(0.043)  & 31.32(0.026) & \textbf{62.86}(\underline{0.012}) & 2.49(0.043)  & 29.86(0.028) & \textbf{56.66}(\underline{0.016})   \\ \hline
\end{tabular*}}

\caption{The original train/test accuracy and loss of model trained with synthesized CIFAR-10/100 images from scratch. The bold type is the best accuracy of the same architecture experiment, and the underbar type is the lowest loss.}
\label{table6}
\end{table*}

\subsection{Feature Visualization on Various Embedding}
We further expand the t-SNE result in feature visualization experiment in Section 4.1. For verifying the feature representation, We choose the feature embedding from each residual block output and embedding layer of ResNet-34. Fig.\ref{tSNE_all} is a visualization of the feature representation of DI and NaturalInverison. Since DI optimizes the input space without non-linearity, it is difficult to distinguish the complex feature of the CIFAR-10 target classes. Therefore, the feature space shows a different distribution from the original distribution. In contrast, our NaturalInversion brings a non-linearity to optimization and causes ease of optimization step with a one-to-one approach generator. Therefore it leads to classify the characteristics of the original target class easily. As a result, our method can represent the original dataset more precisely.

\subsection{From Scratch Training}
We expand the from scratch experiments in Section 4.1 to measure the CIFAR-10/100 top-1 test accuracy of the model trained with synthesized images in Table.\ref{table6}. Also, we calculate the training/test loss by forwarding the CIFAR-10/100 training/test sets to ensure that NaturalInversion smoothes the optimization problem with non-linearity properties.
For experiments, we train the randomly initialized networks from scratch using synthesized images for this study. Then forward original CIFAR-10/100 datasets and calculate the training/test accuracies and loss. Table.\ref{table6} shows the result of training/test accuracies and loss, and we can demonstrate three properties: (a) Our method can capture original training data distribution because the CIFAR-10/100 training accuracy of the model is much higher than others. (b) NaturalInversion converges safely and quickly at optimal points, so the training loss is lower than DAFL and DI. (c) Our method can synthesize generalized images that are less biased to the inverted model, as seen from the test accuracy results. From the results, we can observe that NaturalInversion outperforms prior works on synthesizing more natural images.

\end{document}